\documentclass[runningheads,a4paper]{llncs}
\usepackage{siunitx}

\usepackage{xspace}
\usepackage{amssymb}

\usepackage{graphicx}
\usepackage{amsmath}

\usepackage{algorithm}
\usepackage[noend]{algpseudocode}

\usepackage{array}
\usepackage{color, colortbl}

\makeatletter
\renewcommand{\ALG@beginalgorithmic}{\small}
\makeatother

\usepackage{url}

\definecolor{lightgray}{gray}{0.95}

\begin{document}

\mainmatter  % start of an individual contribution

%\title{Multiple Organ Point Cloud Network for prediction on Medical Shape Data}
\title{ Deep Multi-Structural Shape Analysis: Application to Neuroanatomy}
\titlerunning{Deep Multi-Structural Shape Analysis: Application to Neuroanatomy}

%
%\author{No Author given}

\author{Benjam\'in Guti\'errez-Becker \and
	    Christian Wachinger  
\authorrunning{Benjam\'in Guti\'errez-Becker,  Christian Wachinger} 
\institute{Artificial Intelligence in Medical Imaging (AI-Med), KJP, LMU M\"unchen, Germany}}

\toctitle{Lecture Notes in Computer Science}
\tocauthor{Authors' Instructions}
\maketitle
\footnotetext{Accepted at MICCAI 2018.}

\begin{abstract}
We propose a deep neural network for supervised learning on neuroanatomical shapes. 
The network directly operates on raw point clouds without the need for mesh processing or the identification of point correspondences, as spatial transformer networks map the data to a canonical space. 
Instead of relying on \emph{hand-crafted} shape descriptors, an optimal representation is learned in the end-to-end training stage of the network. The proposed network consists of multiple branches, so that features for multiple structures are learned simultaneously. 
We demonstrate the performance of our method on two applications: (i) the prediction of Alzheimer's disease and mild cognitive impairment and (ii) the regression of the brain age. 
Finally, we visualize the important parts of the anatomy for the prediction by adapting the occlusion method to point clouds. 
%Shape representations are a powerful tool in medical image analysis for its ability to quantify morphological differences. Current approaches are mostly based on \emph{hand-engineered} representations which are potentially not optimal for specific tasks. In this work we propose the use of a neural network operating directly on point cloud representations to perform shape analysis. Our method is able to perform a joint shape analysis of different structures and since it operates directly on raw point cloud representations does not require feature engineering, image registration, mesh calculations or point correspondences. We showcase the advantages of our method by performing prediction based on the analysis of shape of different brain subcortical structures to perform Alzheimer's disease and Mild Cognitive Impairment classification as well as age regression. 
\end{abstract}

\section{Introduction}

% What is shape analysis
Shape analysis of anatomical structures is of core importance for many tasks in medical imaging, not only as a regularization prior for segmentation tasks, but also as a powerful tool to assess differences between subjects and populations.  A fundamental question when operating on shapes is to find a suitable numerical representation for a given task. %\emph{of shape which is general enough to describe all possible variations of shape, but also specific enough to assess differences between subjects}. 
Hence, many different types of parameterizations have been proposed in the past including  point distribution models \cite{Cootes1995}, spectral signatures~\cite{Wachinger2015}, spherical harmonics \cite{gerardin2009multidimensional}, medial representations \cite{Gorczowski2007}, and diffeomorphisms \cite{miller2014diffeomorphometry}.  Even though these representations have proven their utility for the analysis of shapes in the medical domain, they might not be  optimal for a particular task. 
%\emph{hand-engineered} features, which are not necessarily optimal for a particular task. 

In recent years, deep networks have had ample success for many medical imaging tasks by learning complex, hierarchical feature representations from images. These representations have proven to outperform \emph{hand-crafted} features in a variety of medical imaging applications \cite{Litjens2017}. One of the main reasons for the success of these methods is the use of convolutional layers, which take advantage of the shift-invariance properties of images \cite{Bronstein2017}. However, the use of deep networks in medical shape analysis is still largely unexplored; mainly because  typical shape representations such as point clouds and meshes do not possess an underlying Euclidean or grid-like structure. 

In this work, we propose an alternative approach to perform supervised learning on  medical shape data. Our method is based on PointNet~\cite{Qi2017}, a deep neural network architecture, which operates directly on a point cloud and predicts a label in an end-to-end fashion. Point clouds present a raw and simple parameterization that avoids complexities involved with meshes and that is trivial to obtain given a segmented surface. 
The network does not require the alignment of point clouds, as a spatial transformer network maps the data to a canonical space before further processing. 
%PointNet is invariant to the initial alignment of the shapes and does therefore not require shape .
%to permutations of the points by applying max pooling on the learned features across points. 
%The network is able to compute a global descriptor for the entire shape that is used in the classification stage of the network. 
PointNet has been proposed for object classification, where the category of a single shape is predicted. 
For many medical applications however, not just a single anatomical structure is important for the prediction but a simultaneous view of multiple structures is required for a more comprehensive analysis of a subject's anatomy. Hence, we propose the Multi-Structure PointNet (MSPNet), which is able to  simultaneously predict a label given the shape of multiple structures. We evaluate MSPNet in two neuroimaging applications, neurodegenerative disease prediction and age regression.% We modify the standard architecture of PointNet to avoid reducing the learned representation to a global descriptor, but keep localized information that is regularized with dropout.  %but 

\subsection{Related Work}

%Several studies have previously proposed the use of features extracted from shape representations to perform supervised learning tasks. 
Several shape representations have previously been used for supervised learning tasks. % have previously proposed the use of features extracted from shape representations to perform supervised learning tasks. 
Spherical harmonics for approximating the hippocampal shape have been proposed in \cite{gerardin2009multidimensional}.
Shape information has been derived from thickness measurements of the hippocampus from a medial representation~\cite{costafreda2011automated}. 
Statistical shape models to detect hippocampal shape changes were proposed by \cite{shen2012detecting}. 
Multi-resolution shape features with non-Euclidean wavelets were employed for the analysis of cortical thickness \cite{Kim2014107}.  
The use of medial axis shape representations was used to compare the brain morphology of autistic and normal populations \cite{Gorczowski2007}.
Recently, shape representation based on spectral signatures have been introduced to perform age regression and disease prediction~\cite{Wachinger2015,wachinger2016domain}.

All the mentioned approaches rely on computing pre-defined shape features. Alternatively, a variational auto-encoder was proposed to automatically extract features from 3D surfaces, which can in turn be used in a classification task~\cite{Shakeri2016}. However different to our approach, this is not an end-to-end learning since the variational encoder is not directly linked to the classification task. Consequently, the learned features capture overall variation but are not directly optimized for the given task. In addition, this approach relies on computing point correspondences between meshes and focuses on a single structure, while we simultaneously model multiple structures.

\def\real{{\mathbb{R}}}

\def\ourmethod{{MSPNet}}
% Notice that in our case each point is represented not only by its spatial coordinates, but also by its gray-level intensity $g$.
\section{Method}
\begin{figure*}[t]
	\centering\includegraphics[width=0.9\textwidth]{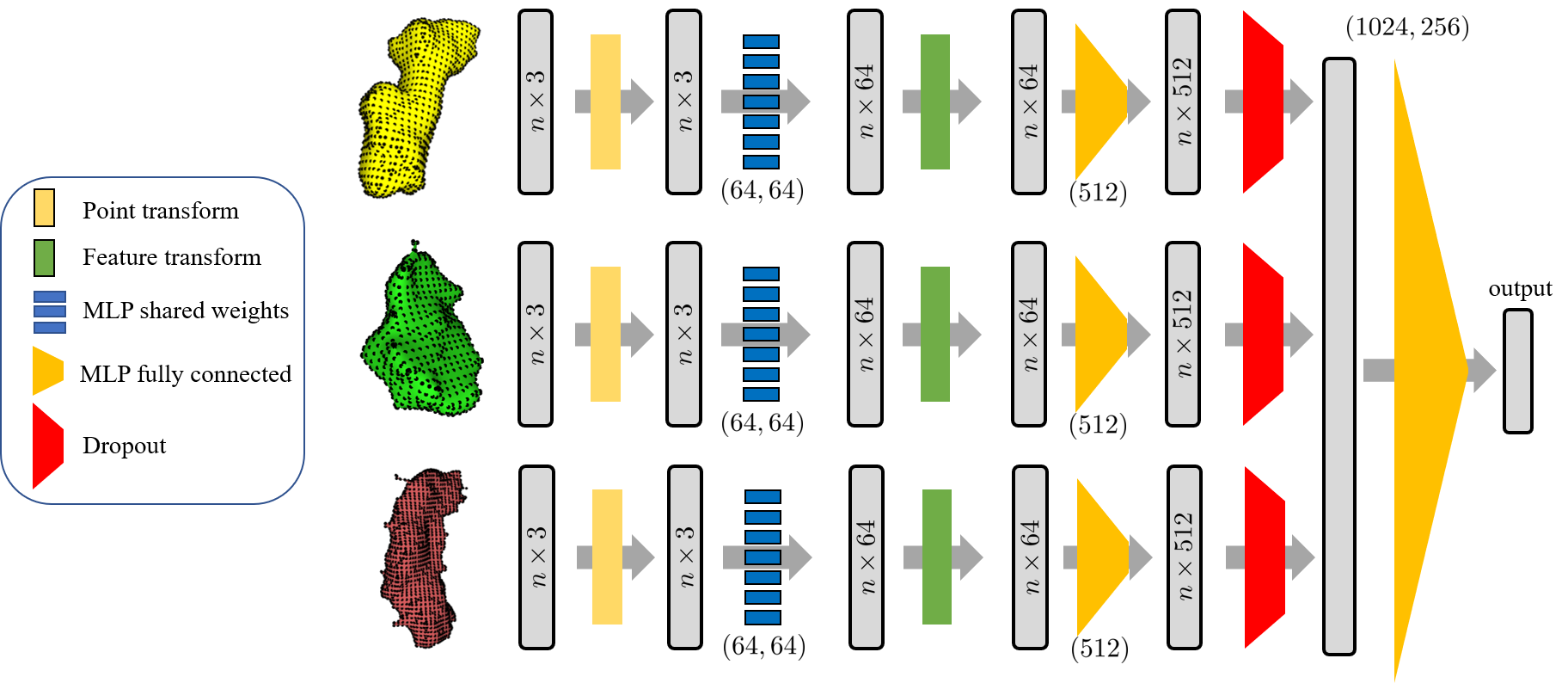}
	\caption{MSPNet Architecture. The network consists of one branch per structure (illustrated for three structures), which are fused before the final  multilayer perceptron (MLP). Each structure is represented by a point cloud with $n$ points that pass through transformer networks and multilayer perceptrons of the individual branch. Numbers in brackets are layer sizes. %Batchnorm is used for all layers with ReLU. Dropout layers are used for the last MLP.
    \label{fig:network}}
\end{figure*}

We propose a method for multiple structure shape analysis that is divided into two main stages: the extraction of point clouds representing the anatomy of different structures from medical images (section \ref{sec:pointcloud}), and a Multi-Structure PointNet (\ourmethod) (section \ref{sec:architecture}). Figure~\ref{fig:network} illustrates the architecture of MSPNet, which is based on   PointNet~\cite{Qi2017}, and extends on it to allow the simultaneous processing of multiple structures. 

%\ourmethod takes the coordinates of unordered sets of points representing the anatomy and predicts an outcome label. 

\noindent
\subsection{Point Cloud Extraction} \label{sec:pointcloud}
We extract point clouds from MRI T1-weighted images of the brain. We process the images with the FreeSurfer pipeline~\cite{Fischl2012} and obtain segmentations of multiple neuroanatomical regions. From the resulting segmentations, point clouds are created by uniformly sampling the boundary of each brain structure. After this process, the anatomy of a subject is represented by a collection  of $m$ point clouds $S=\{ P_0, P_1, \hdots P_m \}$, where each point cloud represents a structure. A point cloud is defined as a set of $n$ points $P = [\mathbf{p}_0, \mathbf{p}_1, \hdots , \mathbf{p}_n] $, where each point is a vector of Cartesian coordinates $\mathbf{p}_i = (x,y,z)$.

\noindent
\subsection{MSPNet Architecture}\label{sec:architecture}
We aim at finding a network architecture corresponding to a function $f : S \mapsto y$, mapping a collection of shapes described by $S$ to a prediction~$y$. An overview of the network is shown in figure \ref{fig:network}. MSPNet consists of multiple branches, where each branch processes the point cloud of one structure independently. This ensures that an optimal feature representation is learned per structure. At the end, the features of all branches are merged to perform a joint prediction.  Each branch can be divided into the following stages: 1) point cloud alignment using a transformation network, 2) feature extraction, 3) feature alignment with a second transformation net, 4) dropout and 5) prediction.  The first three stages of the architecture of each branch resemble that of a single PointNet architecture. The last two stages are particular to \ourmethod.

\begin{figure*}[t]
	\centering\includegraphics[width=0.8\textwidth]{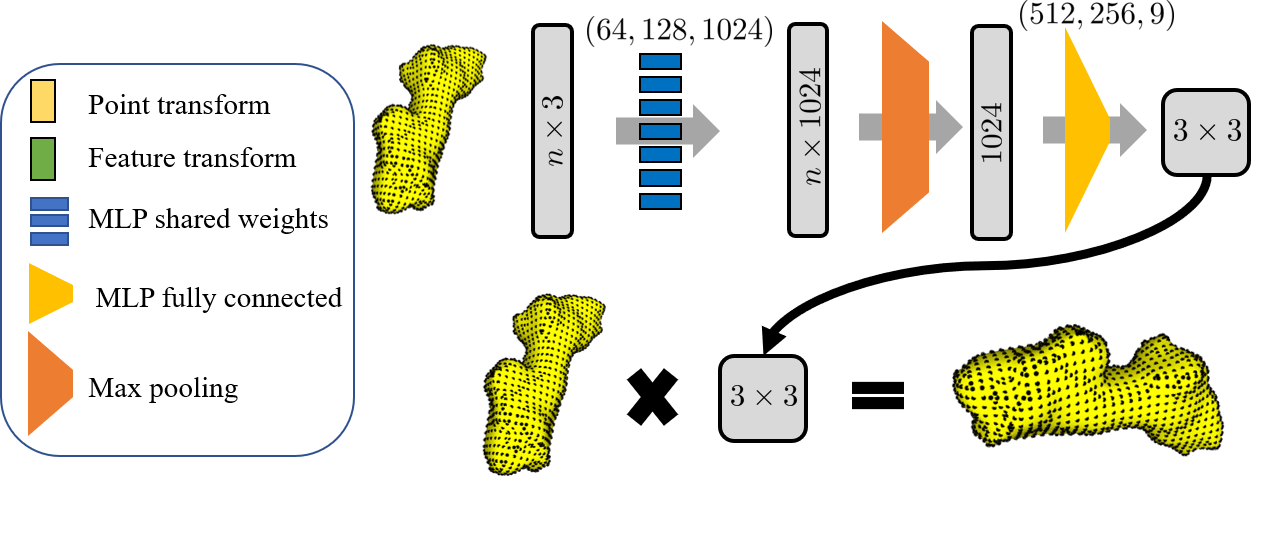}
	\caption{Transformation network (T-Net) for predicting a transformation matrix to map a point cloud to canonical space before processing. A similar network is used to transform the features; the only difference is that the output corresponds to a $64 \times 64$ matrix.	\label{fig:tnet}}
\end{figure*}

\textbf{Point Transformation Network:} In contrast to previous approaches in deep medical shape analysis~\cite{Shakeri2016}, MSPNet does not require point correspondences across shapes, i.e., the i-$th$ points of two shapes, $\mathbf{p}_i^1$ and $\mathbf{p}_i^2$, respectively, do not need to represent the same  anatomical position. We obtain the invariance to rigid transformations in MSPNet by (i) augmenting the training dataset by applying a random rigid transformation to each shape during training time and by (ii) introducing a transformation network (T-Net). This network estimates a $3 \times 3$ transformation matrix, which is applied to the input as a first step. One can think of the T-Net as a transformation into a canonical space to roughly align point clouds before any processing is done.  The T-Net is shown in figure \ref{fig:tnet} and is composed of a multilayer perceptron (MLP), a max pooling operator and two fully connected layers. 

\textbf{Feature Extractions:} The transformed points are fed into a MLP with shared weights among points. This MLP layer can be thought of as the feature extraction stage of the network. At this stage of the network, each point has access to the position of all the remaining points of the point cloud, and therefore as the output of the network, we obtain a $k$-dimensional feature vector for each point (in our case $k$ = 64). Although each point is assigned a single feature vector, in practice each feature vector point contains a global signature of the input point cloud.

\textbf{Feature Transformation:} A second T-Net is applied to the computed features. This network has the same properties as the first transformation network, but its output corresponds to a $k \times k$ transformation matrix. This transformation matrix has a much higher dimension than the previous spatial transformation, which makes the optimization more challenging. To facilitate the optimization of this larger feature transformation matrix $T$, we constrain it to be close to an orthogonal matrix $C_\text{reg} = \| I - T T^\top \|^2_F$, similar to~\cite{Qi2017}. The regularization term ensures a more stable convergence of the network.   After the points are transformed they are fed to a MLP layer.

\textbf{Dropout and prediction:} Up to this point, the architecture of each branch mirrors that of the  PointNet. However the final dropout and prediction stage is particular to \ourmethod. In  PointNet, the last stage corresponds to a max-pooling layer performed across $n$ points, so that the output is a vector with size corresponding to the feature dimensionality. Instead of performing max-pooling, which leads to a strong  shrinkage in feature space, we propose to keep the localized information per point. This leads to an increase in the network capacity, which may lend itself to overfitting. Hence, we introduce a dropout layer (keep probability = 0.3) for regularization. The main advantage of the new design is that more localized information is retained in the network, which we hypothesize may boost the predictive power of our network. Finally, the individual features from each branch are concatenated and fed into a last MLP to perform prediction.  
Batch normalization is used for all MLP layers and ReLU activations are used.  The last MLP perceptron counts with intermediate dropout layers with 0.7 keep probabilities as in PointNet. 
To facilitate the exposition, we assumed that each structure per branch is described by the same number of $n$ points, but in practice each structure can be represented by point clouds of different dimensions.

\noindent

%The other challenge of using a point cloud as input to the network is to ensure that the network response is invariant to the ordering of the points. In PointNet, a symmetric function in the form a max pooling layer is introduced in the network to induce order invariance. However, this order invariance comes at the cost of losing the information given by all the points of the network. Instead of enforcing order invariance with a symmetric function we introduce dropout layers after features for each point are extracted. 
%These drop out layers force the network to learn representations which can perform predictions by looking at a small robust to potentially unordered points. 
 
%Invariance to point ordering can be enforced by introducing a symmetric function which is invariant to permutations in the input.

% This symmetric function corresponds to a max pooling layer which takes as input $n$ vectors and outputs a single vector. 

\section{Results}

We evaluate the performance of MSPNet  in two supervised learning tasks, classification and regression. For the classification task, we aim at using shape descriptors to discriminate between healthy controls (HC), and patients diagnosed  with mild cognitive impairment (MCI) or Alzheimer's disease (AD). For the regression task, we perform age estimation of a subject based on shape information. In all our experiments, we compare to the standard PointNet architecture and spectral shape descriptors in BrainPrint~\cite{Wachinger2015}, which achieved high performance in a competition for Alzheimer's disease classification~\cite{wachinger2016domain}. For PointNet, the multi-structure input corresponds to a concatenation of the point clouds of all structures. 
%For all our experiments we perform comparisons with a spectral signature base method ( \emph{BrainPrint} \cite{Wachinger2015} ), and to a standard PointNet architecture. 
We use image data from the Alzheimer's Disease Neuroimaging Initiative (ADNI) database (adni.loni.usc.edu)~\cite{Jack2008}. 
%The ADNI was launched in 2003 as a public-private partnership, led by Principal Investigator Michael W. Weiner, MD. The primary goal of ADNI has been to test whether serial MRI, PET, other biological markers, and clinical and neuropsychological assessment can be combined to measure the progression of MCI and early AD. For up-to-date information, see www.adni-info.org.
We work with a total of 7,974 images (2,423 HC, 978 AD, and 4,625 MCI).

%  \subsection{Implementation Details}
 
% Our network is implemented in TensorFlow. 
%  %To obtain point clouds for the organs of interest all images were processed with the FreeSurfer pipeline~\cite{Fischl2012}, from where segmentations were obtained. From the segmented images, point clouds were obtained by uniformly sampling the boundary of each structure.   
% Training took X hours, prediction of a single subject took Y seconds. 
% For BrainPrint, surface meshes were constructed from the segmentations with marching cubes. 
% Computation of BrainPrint per structure is about 4 seconds. 
% All the experiments were conducted on an NVIDIA Titan Xp GPU with 12GB RAM.
 
%\subsection{ Alzheimer's disease and Mild Cognitive Impairment Classification based on Shape Data}
\subsection{AD and MCI Classification on Shape Data}

 For this experiment, we perform classification based on the shape of the left and right hippocampus and the left and right lateral ventricles, due to their key importance in Alzheimer's disease~\cite{thompson2004mapping}. Each structure is represented by a Pointcloud of 512 points. For our experiments the dataset is split in a training, validation and test set (75\%,15\%,15\%). Splitting is done on a per subject basis, to guarantee that the same subject does not appear in different sets.
Table \ref{table:classification} reports the results of the classification experiment, where we report average classification precision, recall and F1-score.
% The results of our experiments on the classification task are summarized in table \ref{table:classification}. The results presented in table \ref{table:classification} correspond to the average classification precision, recall and f1-score. 
In both classification scenarios, PointNet shows a higher accuracy than BrainPrint, illustrating the potential of learning feature representations. 
Further, MSPNet showed the best performance, highlighting the benefit of individual feature learning in each branch of the network. 
 %Both PointNet and  BrainPrint presented in general lower F1-scores  when compared to our method for the HC-AD classification task. In the case of HC-MCI the accuracy of MSPNet was again the highest one among the compared methods. 

 \begin{table*}
 	\centering
 %	\resizebox{\columnwidth}{!}
  	\caption{Average precision, recall and F1-score for the mild cognitive impairment and Alzheimer's classification experiments.}
 		\begin{tabular}{ l c c c | c c c }
 			\multicolumn{1}{c}{} &\multicolumn{3}{c}{HC-MCI} & \multicolumn{3}{c}{HC-AD} \\
			%\hline
 			 &  \ \ \ Precision& \ \ \ Recall & \ \ \ F1-score &   \ \ \ Precision & \ \ \ Recall  & \ \ \ F1-score    \\ 
 			\rowcolor{lightgray} 
 			BrainPrint    &0.57 & 0.59 & 0.57& 0.76 &  0.77 & 0.78   \\ 
 			PointNet    & 0.60& 0.61 & 0.59& 0.77 &  0.77 &0.78  \\ 
 			\rowcolor{lightgray} 
			\ourmethod  & 0.62 & 0.60 & \textbf{0.61} & 0.78 & 0.79 & \textbf{0.80}       \\ 
			\end{tabular}
 	\label{table:classification}
 \end{table*}

\subsection{Age Prediction on Shape Data}

\begin{figure}[t!]
\centering
\begin{minipage}{.45\textwidth}
	\centering
	\includegraphics[width=\textwidth]{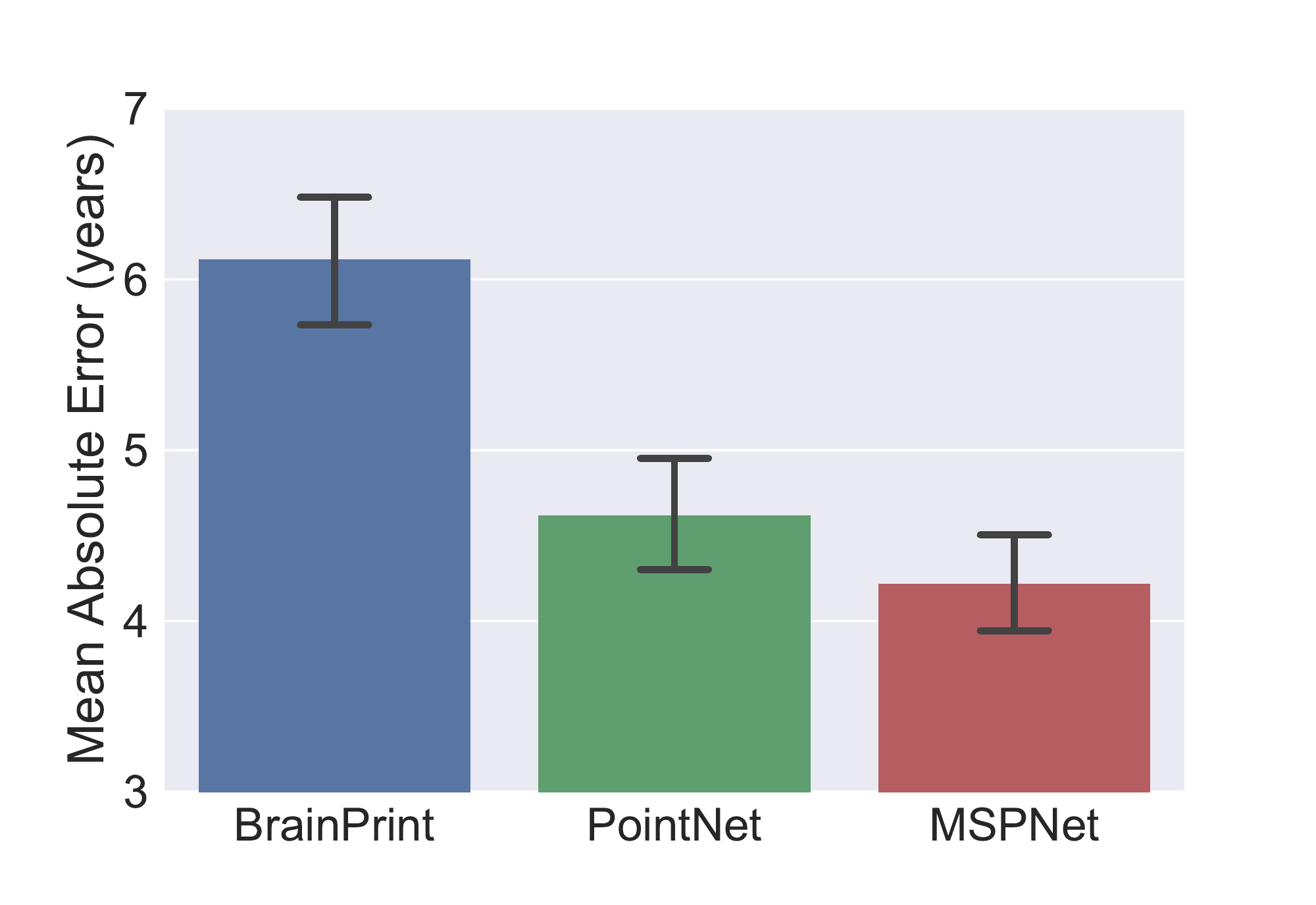}
\end{minipage}
%\hfill
\begin{minipage}{.45\textwidth}
	\centering
	\includegraphics[width=\textwidth]{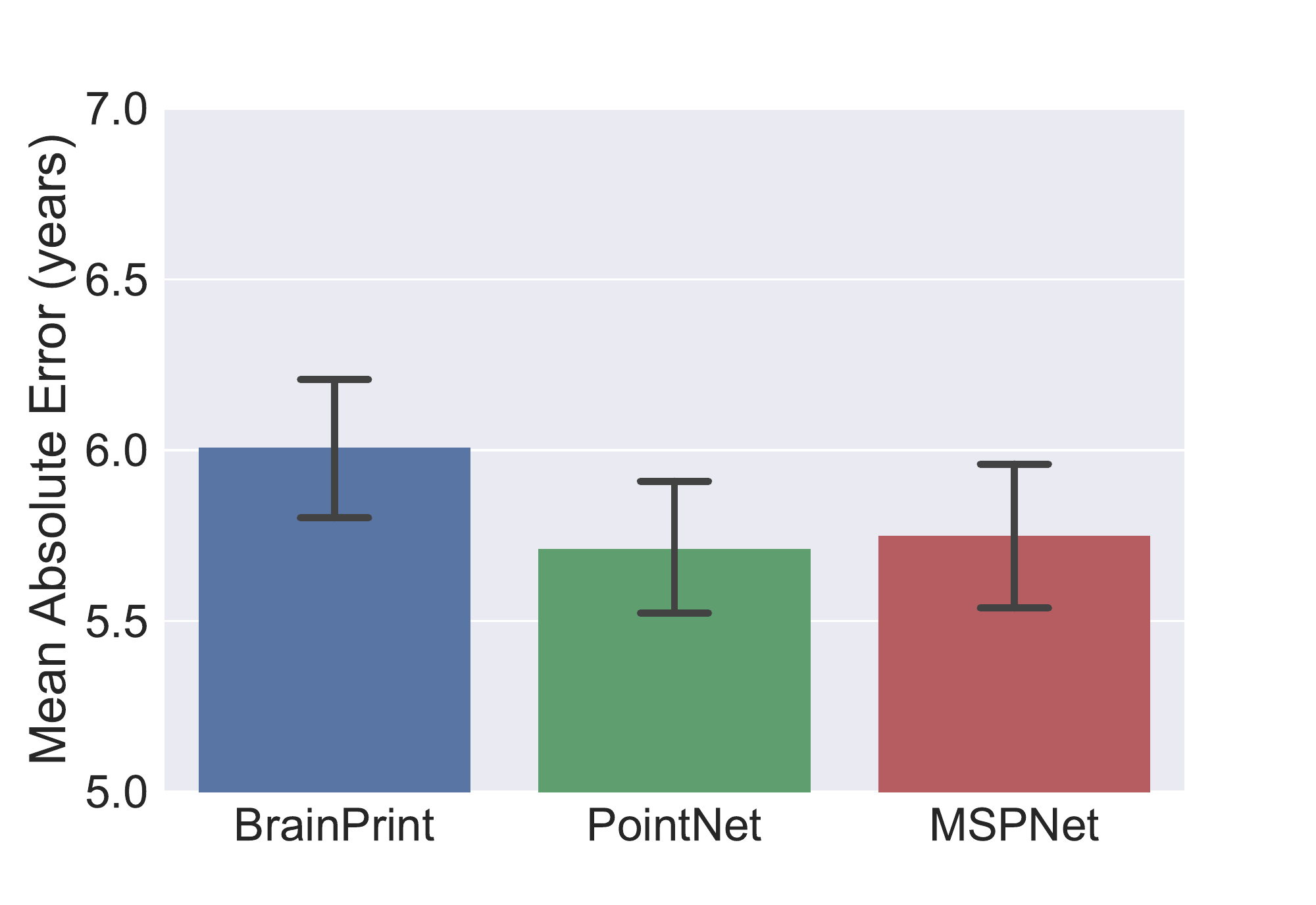}
\end{minipage}
\caption{Mean absolute error for the age prediction experiment on healthy subjects (left) and on all subjects, including MCI and AD (right)}
\label{fig:regression}
\end{figure}

For the age estimation task, we perform two different evaluations. In the first one, we perform age estimation only on the healthy controls of the ADNI database. For the second evaluation, we also include patients diagnosed with MCI and AD. The evaluations are done again on the same brain structures used for the classification task. The results of these two experiments are summarized in the mean absolute error plots of figure \ref{fig:regression}. For the experiment on HC MSPNet significantly outperformed  BrainPrint (p-value \num{2.69e-09}) and PointNet (p-value 0.03). In the experiment on all subjects both PointNet and MSPNet presented comparable performance, both outperforming BrainPrint (p-value 0.01).

\subsection{Visualizing Point Importance}
\begin{figure}[!t]
\centering
\includegraphics[width=0.85\textwidth]{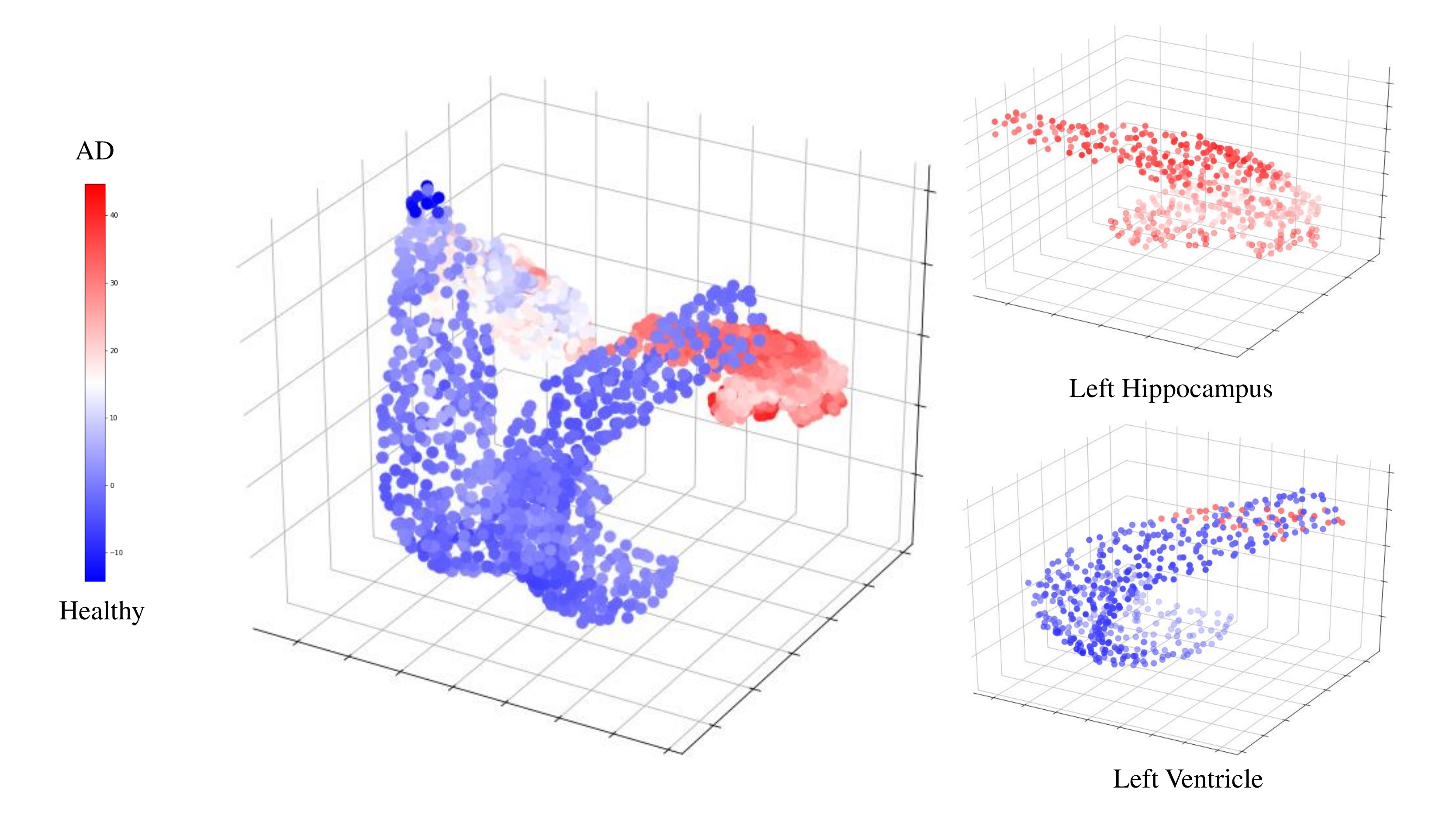}
\caption{Visualization of point importance for the HC-AD classification task for an AD subject. Figure on the left illustrates ventricles and hippocampi, while figures on the right illustrate single structures of the left hemisphere. %Prediction as AD is mainly driven by the shape of the left hippocampus. 
}
\label{fig:visualization}
\end{figure}

Of key importance for making predictions with shapes is the ability to visualize the part of the anatomy that is driving the decision. % made by a classification or a regression algorithm. 
%This is particularly important in the clinical context where shape analysis aids the medical practitioner in the decision making progress. 
This holds in particular in the clinical context. 
In \ourmethod, we introduce a simple yet effective method to visualize the importance that each point has in the prediction. Our visualization is inspired by the commonly used occlusion method \cite{Grun2016}, which consists of occluding parts of a test image and observing differences in the network response. We apply a similar concept to visualize the response of \ourmethod. In our case, we assess the importance of each point in the classification task by occluding this point (making the point coordinates equal to 0) together with its nearest neighbors. Then the occluded point cloud is passed through the network and the response of the output ReLU is compared to that obtained when the full point cloud is evaluated. The difference between these responses can then be assigned as the importance of this particular point. In figure \ref{fig:visualization}, we can observe the result of using this visualization technique for one of the AD test subjects in the HC-AD classification experiment. If a point tends towards the red side of the scale, it indicates that by occluding this particular point, the network increases the activation of the AD class. This means that the region around this point is used by the network to predict AD. The exact opposite is true for points on the blue side of the scale. White points indicate that the network response was not largely affected by occluding this point. In the  particular case of the example in figure \ref{fig:visualization}, the decision of the network to give this subject a AD label is mainly driven by the left hippocampus.  

\section{Conclusion}

We introduced MSPNet, a deep neural network for shape analysis on multiple brain structures. 
%In this work we have presented a method to perform shape analysis on medical data based on point cloud representations. 
To the best of our knowledge, this is the first time that a neural network for shape analysis on point clouds is proposed in medical applications. We have shown that our method is able to achieve high accuracy in both classification and regression tasks, when compared to shape descriptors based on spectral signatures. This performance is achieved without relying on point correspondences or meshes. MSPNet learns feature representations from multiple structures simultaneously. Finally, we illustrated point-wise importance for the prediction by adapting the occlusion method.  \\
\textbf{Acknowledgments.} This work was supported in part by SAP SE and the Bavarian State Ministry of Education, Science and the Arts in the framework of the Centre Digitisation.Bavaria (ZD.B).
\bibliography{biblio}{}
\bibliographystyle{splncs03}

\end{document}